\documentclass[conference]{IEEEtran}
\IEEEoverridecommandlockouts
\usepackage{cite}
\usepackage{amsmath,amssymb,amsfonts}
\usepackage{amsthm}
\usepackage{algorithmic}
\usepackage{graphicx}
\graphicspath{{figures/}}
\usepackage{textcomp}
\usepackage{xcolor}
\usepackage{booktabs}
\usepackage{multirow}
\usepackage[T1]{fontenc}
\usepackage{hyperref}
\hypersetup{
    colorlinks=false,
    pdfborder={0 0 0},
    pdftitle={Ben-Misandry: A Bengali Dataset and Late-Fusion Transformer Framework for Detecting Anti-Male Gender Hate}
}

\def\BibTeX{{\rm B\kern-.05em{\sc i\kern-.025em b}\kern-.08em
    T\kern-.1667em\lower.7ex\hbox{E}\kern-.125emX}}

\def\IEEEtitletopspace{-0.3in}
\begin{document}

\title{%
\makebox[\textwidth][l]{\includegraphics[height=0.45in]{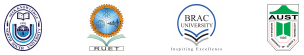}}\\[2pt]
\rule{\textwidth}{1pt}\\[8pt]
{\fontsize{20}{24}\selectfont Ben-Misandry: A Bengali Dataset and Late-Fusion Transformer Framework for Detecting Anti-Male Gender Hate}}

\author{
\IEEEauthorblockN{
Md Mehedi Hasan Nipu\textsuperscript{1},
Chinmoy Mitra\textsuperscript{2},
Tarannum Ahmed Nowshin\textsuperscript{3},
Israt Moyeen Noumi\textsuperscript{4}
}
\IEEEauthorblockA{
\textsuperscript{1}Department of Electrical and Computer Engineering, North South University, Dhaka, Bangladesh\\
\textsuperscript{2}Department of Computer Science \& Engineering, Rajshahi University of Engineering \& Technology, Rajshahi, Bangladesh\\
\textsuperscript{3}Department of Computer Science and Engineering, BRAC University, Dhaka, Bangladesh\\
\textsuperscript{4}Department of Computer Science and Engineering, Ahsanullah University of Science and Technology, Dhaka, Bangladesh\\
mehedi.nipu@northsouth.edu,\; rudrochinmoy@gmail.com,\; taranowshin@gmail.com,\; isratmoyeen.23@gmail.com
}
}

\IEEEaftertitletext{\vspace{-1.2\baselineskip}}
\maketitle

\begin{abstract}
Gender based hate speech research in Bengali has focused almost exclusively on misogyny, leaving anti male hate, or misandry, entirely unstudied for this language and creating a systematic blind spot in moderation systems trained to catch attacks on women but not on men. We introduce Ben-Misandry, the first Bengali misandry detection dataset: 5,000 comments across six taxonomy classes, including a novel category for mocking male emotional vulnerability with no direct analogue in existing misogyny taxonomies. The corpus is taxonomy-controlled synthetic data generated with GPT-4.1, spot-checked by four human reviewers on a 500-comment sample who confirmed the large majority of assigned labels; a full inter-annotator-agreement study is left to future work. We benchmark it with a classical pipeline (seven classifiers over four text representations) and a transformer pipeline (BanglaBERT, XLM-RoBERTa, and two fusion strategies). The best classical model, a linear SVM over character-level TF-IDF, reaches 91.8\% accuracy; a late fusion ensemble averaging BanglaBERT and XLM-RoBERTa softmax outputs reaches 96.6\% accuracy and macro ROC-AUC above 99\%, nominally ahead of every single model and early fusion, though we have not run significance testing to confirm these margins are meaningful. We relate this to classifier combination theory, under which averaging encoders with decorrelated errors is expected to reduce variance, and corroborate the decision boundary with Shapley explanations that recover the same class-discriminative vocabulary found in our exploratory analysis, while noting this cannot rule out the model partly keying on synthetic-generation artifacts. Our results indicate misandry detection in Bengali is feasible and measurable on this synthetic corpus, offered as an initial public resource pending validation on real-world data.
\end{abstract}

\begin{IEEEkeywords}
misandry detection, Bengali natural language processing, gender based hate speech, transformer fusion, BanglaBERT, XLM-RoBERTa, explainable artificial intelligence
\end{IEEEkeywords}

\section{Introduction}

Automated moderation for gender based hate has almost always been built in one direction: hostility toward women. Bengali corpora such as Ben-Misog \cite{b1}, BOISHOMMO \cite{b2}, and Karim et al. \cite{b3} label anti-woman or general gender-tagged hostility, but none label the opposite direction, and misandry is nearly absent even from broad hate-speech surveys \cite{b22}; a search of HuggingFace Datasets and public GitHub repositories for a Bengali misandry corpus returns no results. Moderation classifiers therefore cannot recognize anti-male hostility even when the same tropes, contempt, stereotypes, threats, dismissal of suffering, are used against men almost word for word. This paper builds the missing labeled resource, mirroring the peer reviewed Ben-Misog taxonomy \cite{b1} for future cross-comparison while adding one category with no misogyny-taxonomy counterpart: mockery and emotional dismissal, the pattern of ridiculing male vulnerability and treating male suffering as illegitimate.

Our contributions are: (1) \textbf{Ben-Misandry}, the first Bengali misandry dataset, five thousand comments balanced across six classes and checked through log-odds, male-referent statistics, and a 500-row human spot-check; (2) a classical benchmark spanning seven classifiers and four text representations (twenty five configurations); (3) a transformer benchmark comparing BanglaBERT, XLM-RoBERTa, and early/late fusion with two late-fusion rules; (4) a discussion, grounded in classifier combination theory \cite{b7,b8,b9}, of why late fusion of two structurally different encoders is expected to outperform either alone; and (5) a Shapley-based explainability analysis \cite{b10} that recovers vocabulary consistent with our corpus statistics, discussed alongside the risk of lexical shortcuts.

\section{Related Work}

\subsection{Bengali gender hate detection}
Roy et al. \cite{b1} introduced Ben-Misog, the misogyny focused Bengali corpus and six class taxonomy this work mirrors. BOISHOMMO \cite{b2} and BanTH \cite{b21} are large multi-label Bengali hate corpora with a gender category but general, non-directional targets. Karim et al. \cite{b3} released a Bengali hate speech dataset and DeepHateExplainer, an explainable CNN-LSTM/transformer pipeline. None annotate anti-male content as a first-class category, the gap Ben-Misandry closes.

\subsection{Transformer based Bengali text classification}
BanglaBERT \cite{b4} is an ELECTRA-style encoder \cite{b12} pretrained on 27.5GB of Bengali text and the strongest public monolingual Bengali encoder for classification; it and BERT \cite{b14} share the same masked/discriminative pretraining lineage. XLM-RoBERTa \cite{b5} is a multilingual RoBERTa-style encoder \cite{b13} pretrained on one hundred languages, lacking Bengali specialization but benefiting from cross-lingual transfer. Both have been used individually for Bengali hate speech, but not fused, and not on a gender hate task.

\subsection{Classifier combination theory}
Kittler et al. \cite{b7} showed the sum rule (probability averaging) is more robust to posterior estimation error than product or maximum rules. Krogh and Vedelsby \cite{b8} and Ueda and Nakano \cite{b9} derived bias-variance-covariance decompositions showing ensemble error is bounded by average individual error minus a diversity term. We apply these results in Section IV to justify late fusion.

\section{The Ben-Misandry Dataset}

\subsection{Taxonomy}
Table \ref{tab:taxonomy} summarizes the six class taxonomy used to label Ben-Misandry, locked before any data was generated. Each class carries a negative rubric excluding criticism of patriarchy as a system, reports of a specific male-perpetrated crime, and feminist advocacy from the misandric classes, so the dataset penalizes gender based contempt specifically, not any mention of men.

\begin{table}[htbp]
\caption{Ben-Misandry taxonomy}
\begin{center}
\footnotesize
\begin{tabular}{@{}p{0.26\linewidth}p{0.62\linewidth}@{}}
\toprule
\textbf{Class} & \textbf{Definition} \\
\midrule
Discredit & Belittles men's competence or worth as a gender. \\
Stereotype \& Objectification & Reduces men to a provider, emotionless, or aggressive trope. \\
Mockery \& Emotional Dismissal\textsuperscript{*} & Mocks male emotional expression or dismisses male suffering. \\
Threats \& Violence & Explicit or implicit threats, or endorsement of violence, against men. \\
Dominance \& Derailing & Female-supremacist framing, or derailing discussion of men's issues. \\
Non-misandric & Neutral, factual, or pro-equality content; control class. \\
\bottomrule
\end{tabular}
\end{center}
\footnotesize\textsuperscript{*}Novel category with no analogue in existing Bengali misogyny taxonomies.
\label{tab:taxonomy}
\end{table}

\subsection{Construction}
Genuine anti-male Bengali content is rare online and hard to mine ethically at scale, so Ben-Misandry v1 is a taxonomy-controlled synthetic corpus generated with GPT-4.1 under a research-framed prompt, followed by deterministic cleanup. Generation used four-shot prompting per category with randomized register, length, and male referent choice (\textit{purush}, \textit{chele}, \textit{shami}, \textit{baba}, \textit{bhai}, \textit{purushmanush}, \textit{betachhele}) per batch, at temperature 0.9, to avoid template collapse. Roughly twenty percent of the initial generation contained banglish or English loanwords; a three-pass transliteration procedure and a small substitution table brought residual Latin-character rate to zero, and exact-text deduplication removed zero additional rows. The release contains five thousand rows (\texttt{text}, \texttt{class}), fully Bengali script, imbalance ratio 1.002.

All rows are synthetic, generated and labeled by construction rather than mined and annotated from real social media. As a first check on label quality, four native-Bengali reviewers manually inspected a stratified random sample of 500 rows (roughly 83 per class), split into non-overlapping quarters of about 125 rows per reviewer, and judged whether each comment's assigned class matched the taxonomy in Table \ref{tab:taxonomy}. The large majority of sampled labels were confirmed correct, with occasional flags concentrated in the same conversational classes discussed in Section VI. Because each row was seen by exactly one reviewer, this supports single-rater label plausibility but does not yield an inter-annotator agreement statistic; a fully overlapping multi-annotator pass with Cohen's $\kappa$/Krippendorff's $\alpha$ reporting is left to future work (Section VII).

\subsection{Corpus statistics and validation}
Table \ref{tab:stats} summarizes corpus level statistics. The dataset averages 10.2 words and 58.9 characters per comment; 13.7\% of rows contain at least one emoji, most concentrated in \texttt{mockery\_emotional\_dismissal}, consistent with its sarcastic laughing tone.

\begin{table}[htbp]
\caption{Ben-Misandry corpus statistics}
\begin{center}
\footnotesize
\begin{tabular}{@{}lr@{}}
\toprule
\textbf{Statistic} & \textbf{Value} \\
\midrule
Total comments & 5{,}000 \\
Classes & 6 (imbalance ratio 1.002) \\
Unique texts & 5{,}000 (0\% duplicates) \\
Bengali script coverage & 100\% \\
Rows containing any emoji & 683 (13.7\%) \\
Mean words / comment & 10.2 (median 10.0) \\
Global unique vocabulary & 4{,}063 word types \\
Total token occurrences & 50{,}425 \\
\bottomrule
\end{tabular}
\end{center}
\label{tab:stats}
\end{table}

Because the six classes are near-uniformly sized (833--835 rows each, imbalance ratio 1.002), accuracy and macro F1 are unlikely to be dominated by a single majority class; per-class results in Table \ref{tab:perclass} confirm every class, not just the largest, is learned to a comparable standard.

\begin{figure}[htbp]
\centerline{\includegraphics[width=0.8\linewidth]{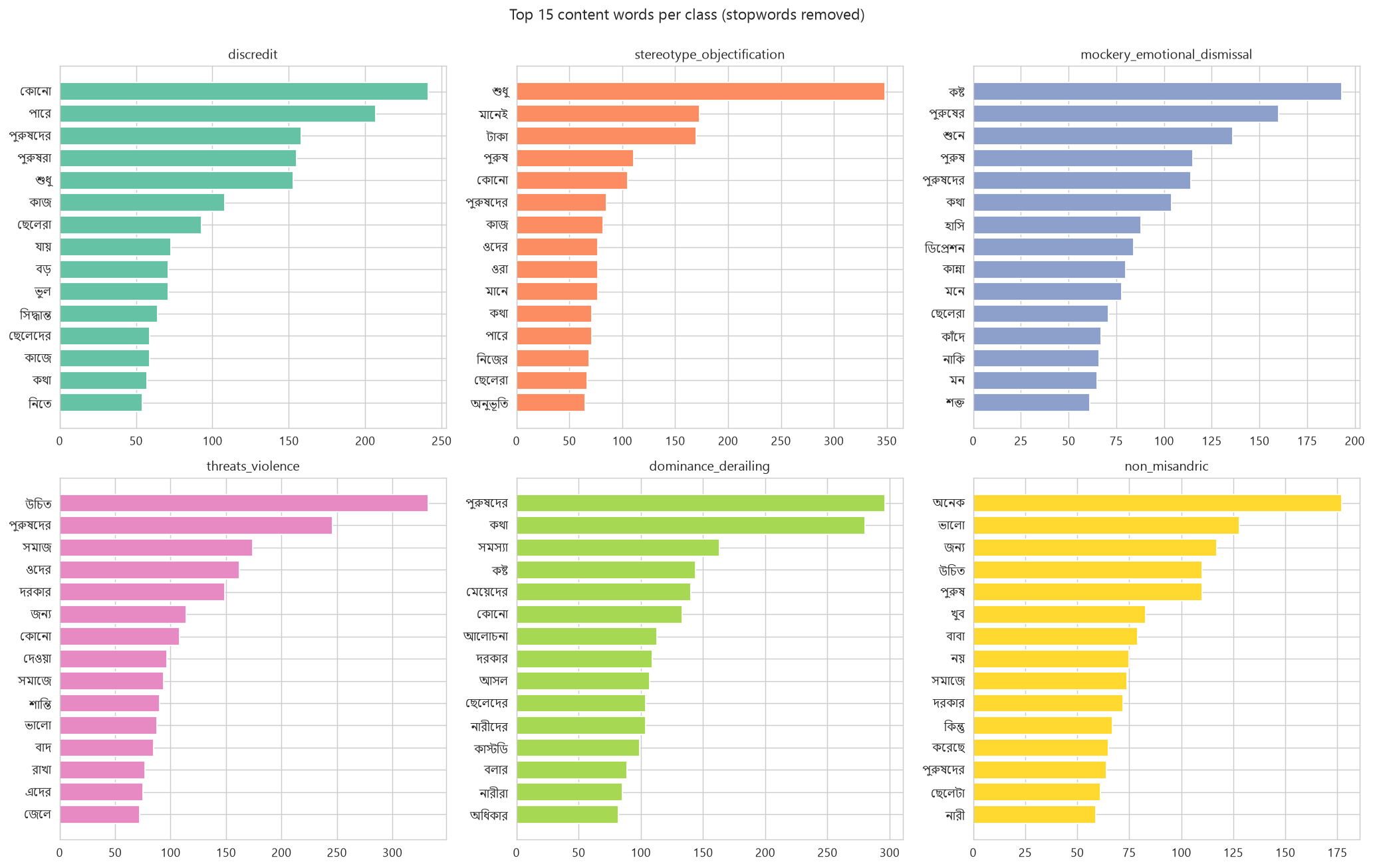}}
\caption{Top fifteen content words per class after stopword removal. Vocabulary is visibly disjoint: discredit centers on capability words, stereotype/objectification on reductive nouns, mockery/emotional dismissal on emotion and suffering words.}
\label{fig:topwords}
\end{figure}

\begin{figure}[htbp]
\centerline{\includegraphics[width=0.8\linewidth]{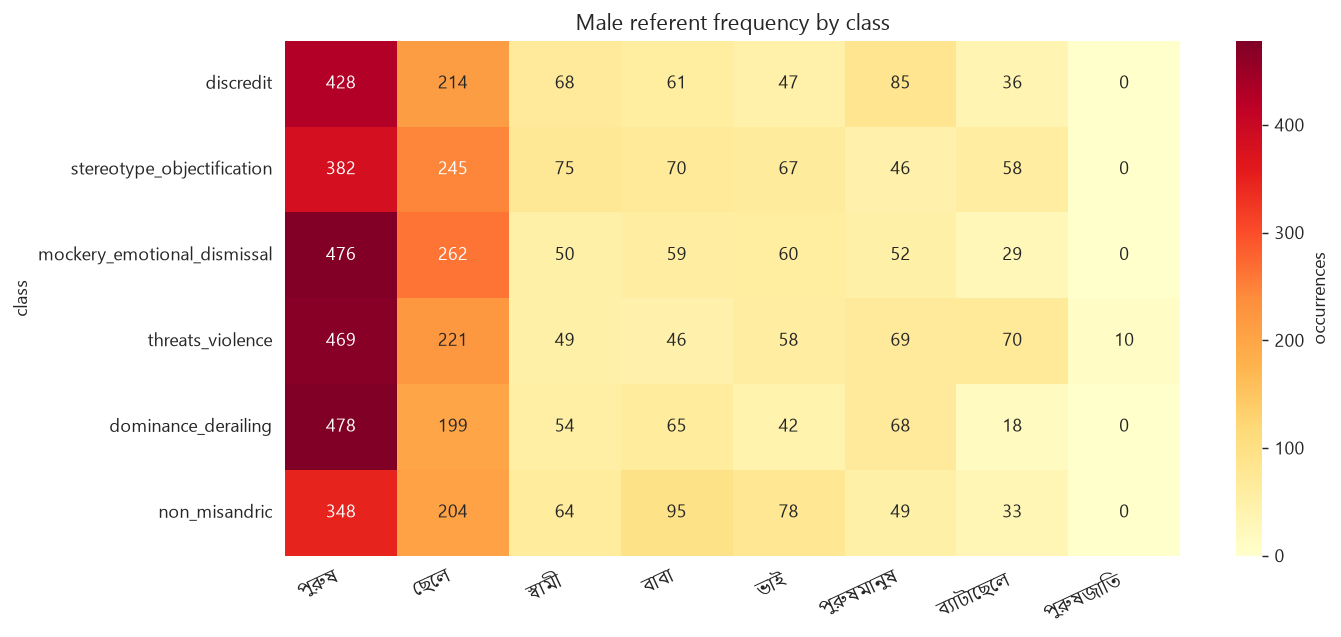}}
\caption{Frequency of male referent tokens by class. Every class, including the control, is dense in male referents. \textit{Purushjati} (male race) occurs only in threats\_violence.}
\label{fig:malereferent}
\end{figure}

As a construct-validity check, we compute the log-odds ratio of every content word against the remaining five classes. The highest positive log-odds words match the intended semantics: \textit{kastodi} (custody, $+10.63$) and \textit{narirai} (women alone, $+9.90$) dominate \texttt{dominance\_derailing}; \textit{jele} (in jail, $+10.49$) and \textit{boycott} ($+9.59$) dominate \texttt{threats\_violence}; \textit{comedy} ($+9.31$) and \textit{nyakami} (mock-crying, $+9.10$) dominate \texttt{mockery\_emotional\_dismissal}. Fig.~\ref{fig:topwords} shows the top fifteen content words per class, and Fig.~\ref{fig:malereferent} shows every misandric class is dense in male referent tokens (871 to 992 occurrences per roughly 833 rows), confirming the corpus is consistently about men rather than incidentally mentioning them.

\section{Methodology}

\subsection{Preprocessing}
Both pipelines apply the BUET Bengali normalizer \cite{b4}, matching BanglaBERT's pretraining convention. The classical pipeline additionally strips URLs, handles, hashtags, digits, and punctuation, and removes roughly eighty Bengali stopwords, since bag-of-words and TF-IDF are more sensitive to noisy tokens than a contextual encoder. Gender referent tokens are retained in both pipelines as the most discriminative feature family (Fig.~\ref{fig:malereferent}).

\subsection{Data splitting and class balancing}
Exact-text deduplication was applied to the full 5,000-row corpus before splitting (Section III-B) and removed zero rows, so no exact duplicate exists anywhere in the dataset, including across splits; we did not separately search for near-duplicate paraphrases, so residual paraphrase-level leakage cannot be fully ruled out. Both pipelines use an identical, stratified 72/18/10 train/validation/test split, giving a shared five hundred row held-out test set. Class balancing by random oversampling with replacement is applied to the training split only:
\begin{equation}
n^{*} = \max_{c \in \mathcal{C}} n_c, \qquad
\tilde{D}_c = \operatorname{Resample}(D_c,\, n^{*},\, \text{replace}=\text{True})
\label{eq:oversample}
\end{equation}
where $D_c$ is the set of training rows of class $c$ and $\mathcal{C}$ is the set of six classes. Validation and test splits are never resampled, so reported test metrics reflect the natural class distribution.

\subsection{Classical machine learning pipeline}
We extract four text representations, each fit only on the training split.

\begin{itemize}
\item \emph{Bag of words}: raw unigram counts, $\mathrm{min\_df}=2$, capped at 30{,}000 features.
\item \emph{Word TF-IDF} \cite{b17}: unigram term frequency inverse document frequency,
\begin{equation}
\mathrm{tfidf}(t,d) = \big(1+\log tf(t,d)\big)\cdot \log\frac{N}{df(t)}
\label{eq:tfidf}
\end{equation}
with sublinear scaling, $\mathrm{min\_df}=2$, capped at 30{,}000 features, where $N$ is the number of training documents and $df(t)$ is the number containing term $t$.
\item \emph{Word 1-2 gram TF-IDF}: as \eqref{eq:tfidf} extended to bigrams, capped at 50{,}000 features.
\item \emph{Character TF-IDF}: as \eqref{eq:tfidf} over character 3 to 5-grams, capped at 50{,}000 features. Bengali is morphologically rich, so sub-word spans capture signal whole-word tokenization discards.
\end{itemize}

We train seven classifiers per representation (Gradient Boosting only on the smallest, for tractability): multinomial naive Bayes, logistic regression, linear SVM \cite{b15}, random forest \cite{b16} (300 trees), decision tree, $k$-NN ($k=5$, cosine), and gradient boosting (150 estimators, depth 3). The linear SVM, strongest of the classical models, solves
\begin{equation}
\min_{w,b}\ \frac{1}{2}\lVert w \rVert^2 + C\sum_{i=1}^{n} \max\!\big(0,\, 1 - y_i (w^{\top}x_i + b)\big)
\label{eq:svm}
\end{equation}
per class under a one-versus-rest reduction, where $x_i$ is the sparse feature vector of comment $i$ and $C$ is the regularization strength. Twenty five representation-classifier pairs are trained in total.

\subsection{Transformer pipeline}

\paragraph{Single encoders} We fine-tune \texttt{csebuetnlp/banglabert} \cite{b4} and \texttt{xlm-roberta-base} \cite{b5}, both descendants of the BERT pretraining lineage \cite{b14}, independently as sequence classifiers. Each produces $h \in \mathbb{R}^{768}$ at its pooling position (\texttt{[CLS]} for BanglaBERT, \texttt{<s>} for XLM-RoBERTa), with a linear head $p = \operatorname{softmax}(W h + b)$, $W \in \mathbb{R}^{6\times768}$. Both are fine-tuned with cross-entropy loss, AdamW \cite{b18}, learning rate $2\times10^{-5}$, weight decay 0.01, max sequence length 64, batch size 16, up to 10 epochs with early stopping on validation weighted F1 (patience 2).

\paragraph{Early fusion} Early fusion concatenates the two encoders' pooled representations before classification:
\begin{align}
z &= \big[\, h_{\mathrm{bb}} \, \Vert \, h_{\mathrm{xr}} \,\big] \in \mathbb{R}^{1536} \label{eq:concat}\\
f &= \operatorname{Dropout}_{0.3}\!\big(\operatorname{ReLU}(W_1 z + b_1)\big) \in \mathbb{R}^{512} \label{eq:fusehead}\\
p_{\mathrm{EF}} &= \operatorname{softmax}(W_2 f + b_2) \label{eq:fuseout}
\end{align}
where $h_{\mathrm{bb}}$, $h_{\mathrm{xr}}$ are the two encoders' pooled vectors, tokenized separately. The full model trains end to end with a mixed-precision loop, batch size 16, up to 20 epochs with early stopping.

\paragraph{Late fusion} Late fusion keeps the two independently fine-tuned encoders as separate classifiers and combines their output probabilities at inference. Let $p_{\mathrm{bb}}, p_{\mathrm{xr}} \in \Delta^{5}$ denote the two softmax outputs over six classes. We evaluate two combination rules.

\emph{Averaging (sum rule):}
\begin{equation}
p_{\mathrm{LF}} = \tfrac{1}{2}\big(p_{\mathrm{bb}} + p_{\mathrm{xr}}\big), \qquad \hat{y} = \arg\max_k\, p_{\mathrm{LF},k}
\label{eq:avg}
\end{equation}

\emph{Max-confidence voting:}
\begin{equation}
p_{\mathrm{LF}} =
\begin{cases}
p_{\mathrm{bb}} & \text{if } \max_k p_{\mathrm{bb},k} \ge \max_k p_{\mathrm{xr},k} \\
p_{\mathrm{xr}} & \text{otherwise}
\end{cases}
\label{eq:maxconf}
\end{equation}

Late fusion needs no additional parameters beyond the two independent fine-tuning runs, making it the cheapest fusion strategy.

\subsection{Why might late fusion outperform its components?}

Averaging rule \eqref{eq:avg} is a special case of the classifier combination problem analyzed by Kittler et al. \cite{b7} and the bias-variance-covariance decomposition of Ueda and Nakano \cite{b9}; we apply, rather than claim novelty in, this result. Let $p_{\mathrm{bb},k}(x)$, $p_{\mathrm{xr},k}(x)$ be the two encoders' softmax outputs for class $k$ estimating posterior $P(y=k\mid x)$, with error $e_i(x) = p_{i,k}(x) - P(y=k\mid x)$. Writing $\sigma_i^2 = \mathbb{E}_x[e_i(x)^2]$ and $\rho = \operatorname{Corr}(e_{\mathrm{bb}}, e_{\mathrm{xr}})$, the variance-of-a-sum identity gives
\begin{equation}
\operatorname{Var}(\bar{e}) = \frac{\sigma_{\mathrm{bb}}^2 + \sigma_{\mathrm{xr}}^2 + 2\rho\,\sigma_{\mathrm{bb}}\sigma_{\mathrm{xr}}}{4}.
\label{eq:varavg}
\end{equation}
When $\sigma_{\mathrm{bb}} = \sigma_{\mathrm{xr}} = \sigma$, this reduces to $\sigma^2(1+\rho)/2$, strictly below $\sigma^2$ whenever $\rho<1$, with no benefit only as $\rho \to 1$.

This is a plausibility argument, not empirical proof: it targets squared-error risk on the unobserved true posterior and assumes $\rho<1$ rather than measuring it; we did not compute $\rho$ from per-example outputs, leaving this to future work. Our circumstantial evidence is architectural: the two encoders differ in pretraining objective, corpus, and tokenizer \cite{b12,b13}, and Table \ref{tab:perclass} shows they disagree on their second-weakest class, consistent with, but not proof of, decorrelated errors. We report the observed gains in Section V at face value and do not claim this argument alone establishes late fusion must outperform its components.

\subsection{Evaluation metrics}
We report accuracy, weighted/macro F1, weighted precision and recall, Cohen's kappa \cite{b19}, and the Matthews correlation coefficient \cite{b20}. Weighted F1 weights per-class harmonic mean of precision and recall by class support:
\begin{equation}
F1_w = \sum_{k=1}^{6} \frac{n_k}{N}\cdot \frac{2\,\mathrm{Prec}_k\cdot\mathrm{Rec}_k}{\mathrm{Prec}_k+\mathrm{Rec}_k}
\end{equation}
Cohen's kappa corrects observed agreement $p_o$ for chance agreement $p_e$:
\begin{equation}
\kappa = \frac{p_o - p_e}{1-p_e}
\end{equation}

\section{Results}

\subsection{Overall comparison}
Table \ref{tab:allresults} reports every transformer configuration and the best classical configuration per family, all on the same held-out five hundred row test set; the full twenty five classical configurations appear in our released results file.

\begin{table*}[htbp]
\caption{Test set results, all transformer configurations and the best classical configuration per model family}
\begin{center}
\footnotesize
\begin{tabular}{@{}llrrrrrr@{}}
\toprule
\textbf{Family} & \textbf{Model (best feature)} & \textbf{Acc.} & \textbf{F1$_w$} & \textbf{F1$_m$} & \textbf{Prec.$_w$} & \textbf{Rec.$_w$} & \textbf{$\kappa$} \\
\midrule
Transformer & \textbf{Late Fusion (Averaging)} & \textbf{0.966} & \textbf{0.9660} & \textbf{0.9660} & \textbf{0.9663} & \textbf{0.966} & \textbf{0.9592} \\
Transformer & Late Fusion (Max-Conf. Voting) & 0.966 & 0.9660 & 0.9660 & 0.9663 & 0.966 & 0.9592 \\
Transformer & Early Fusion & 0.960 & 0.9601 & 0.9601 & 0.9607 & 0.960 & 0.9520 \\
Transformer & BanglaBERT (single) & 0.958 & 0.9579 & 0.9579 & 0.9581 & 0.958 & 0.9496 \\
Transformer & XLM-R-base (single) & 0.950 & 0.9500 & 0.9500 & 0.9506 & 0.950 & 0.9400 \\
\midrule
Classical & Linear SVM (char\_wb 3-5) & 0.918 & 0.9181 & 0.9181 & 0.9192 & 0.918 & 0.9016 \\
Classical & Logistic Reg. (char\_wb 3-5) & 0.914 & 0.9139 & 0.9140 & 0.9148 & 0.914 & 0.8968 \\
Classical & Multinomial NB (word 1-2gram) & 0.872 & 0.8719 & 0.8719 & 0.8743 & 0.872 & 0.8464 \\
Classical & Gradient Boosting (word 1gram) & 0.810 & 0.8106 & 0.8108 & 0.8145 & 0.810 & 0.7720 \\
Classical & Random Forest (char\_wb 3-5) & 0.810 & 0.8082 & 0.8084 & 0.8117 & 0.810 & 0.7720 \\
Classical & KNN, $k$=5, cosine (BoW 1gram) & 0.796 & 0.7964 & 0.7964 & 0.7982 & 0.796 & 0.7552 \\
Classical & Decision Tree (word 1-2gram) & 0.656 & 0.6584 & 0.6585 & 0.6828 & 0.656 & 0.5871 \\
\bottomrule
\end{tabular}
\end{center}
\label{tab:allresults}
\end{table*}

Two observations follow. First, character-level sub-word features consistently beat word-level features for the classical family, confirming Bengali's morphological richness carries signal below the word level. Second, every transformer configuration scores above every classical one; the best classical model trails the best transformer by 4.8 accuracy points while training in 0.09s on CPU versus GPU-minutes for fine-tuning. These gaps are point estimates on a single 500-row split; we did not run paired significance testing (e.g. McNemar's test) or bootstrap confidence intervals, so whether the 0.6-point Late-Fusion-vs-best-single-encoder gap is distinguishable from noise remains open, and we flag this as a limitation.

\subsection{Late fusion: detailed results}
Table \ref{tab:perclass} reports the per-class classification report for the best model, Late Fusion by probability averaging.

\begin{table}[htbp]
\caption{Per-class test results, Late Fusion (Averaging)}
\begin{center}
\footnotesize
\begin{tabular}{@{}p{0.34\linewidth}rrrr@{}}
\toprule
\textbf{Class} & \textbf{Prec.} & \textbf{Rec.} & \textbf{F1} & \textbf{Supp.} \\
\midrule
Discredit & 0.9518 & 0.9518 & 0.9518 & 83 \\
Dominance \& Derailing & 0.9405 & 0.9518 & 0.9461 & 83 \\
Mockery \& Emo. Dismissal & 0.9875 & 0.9518 & 0.9693 & 83 \\
Non-misandric & 0.9540 & 0.9881 & 0.9708 & 84 \\
Stereotype \& Objectification & 0.9643 & 0.9643 & 0.9643 & 84 \\
Threats \& Violence & 1.0000 & 0.9880 & 0.9939 & 83 \\
\bottomrule
\end{tabular}
\end{center}
\label{tab:perclass}
\end{table}

\texttt{threats\_violence} is the strongest class for every model, consistent with its lexically distinct vocabulary (\textit{jele} ``in jail'', \textit{boycott}, \textit{nishiddho} ``banned''). The weakest, \texttt{dominance\_derailing} (F1 0.9461), is occasionally confused with \texttt{discredit}, which shares surface vocabulary around men's competence.

\begin{figure}[htbp]
\centerline{\includegraphics[width=0.85\linewidth]{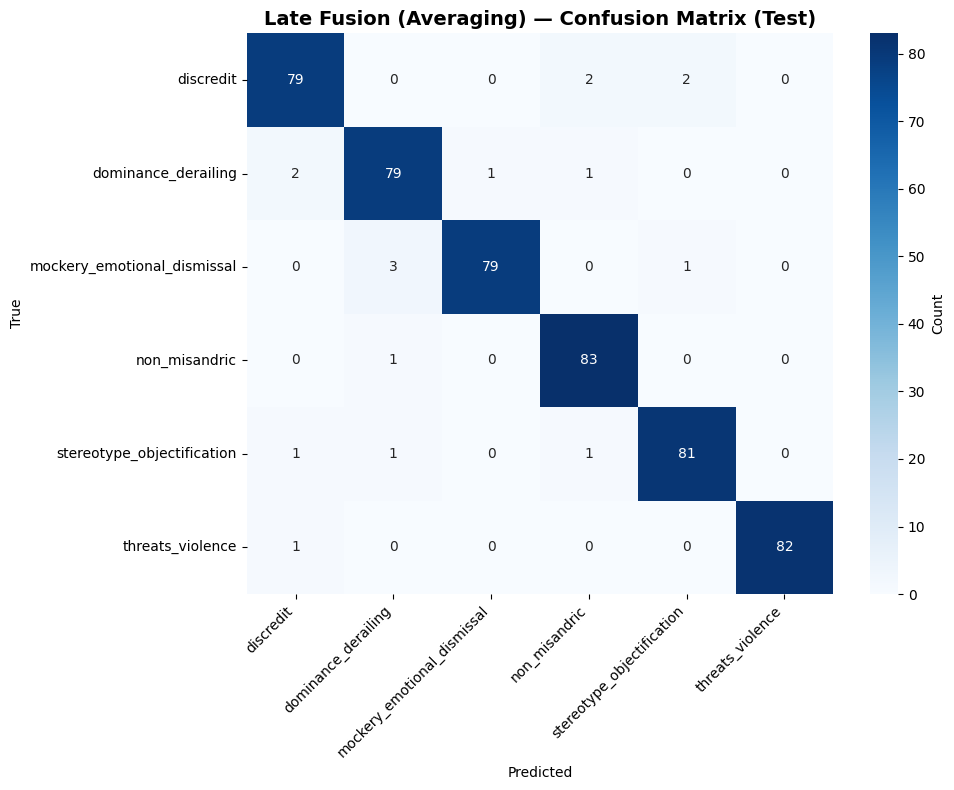}}
\caption{Confusion matrix, Late Fusion (Averaging), held-out test set.}
\label{fig:confmat}
\end{figure}

Fig.~\ref{fig:confmat} shows the confusion matrix: off-diagonal mass is small and diffuse, three or fewer misclassifications per class pair, indicating the six classes are well separated.

\begin{figure}[htbp]
\centerline{\includegraphics[width=0.85\linewidth]{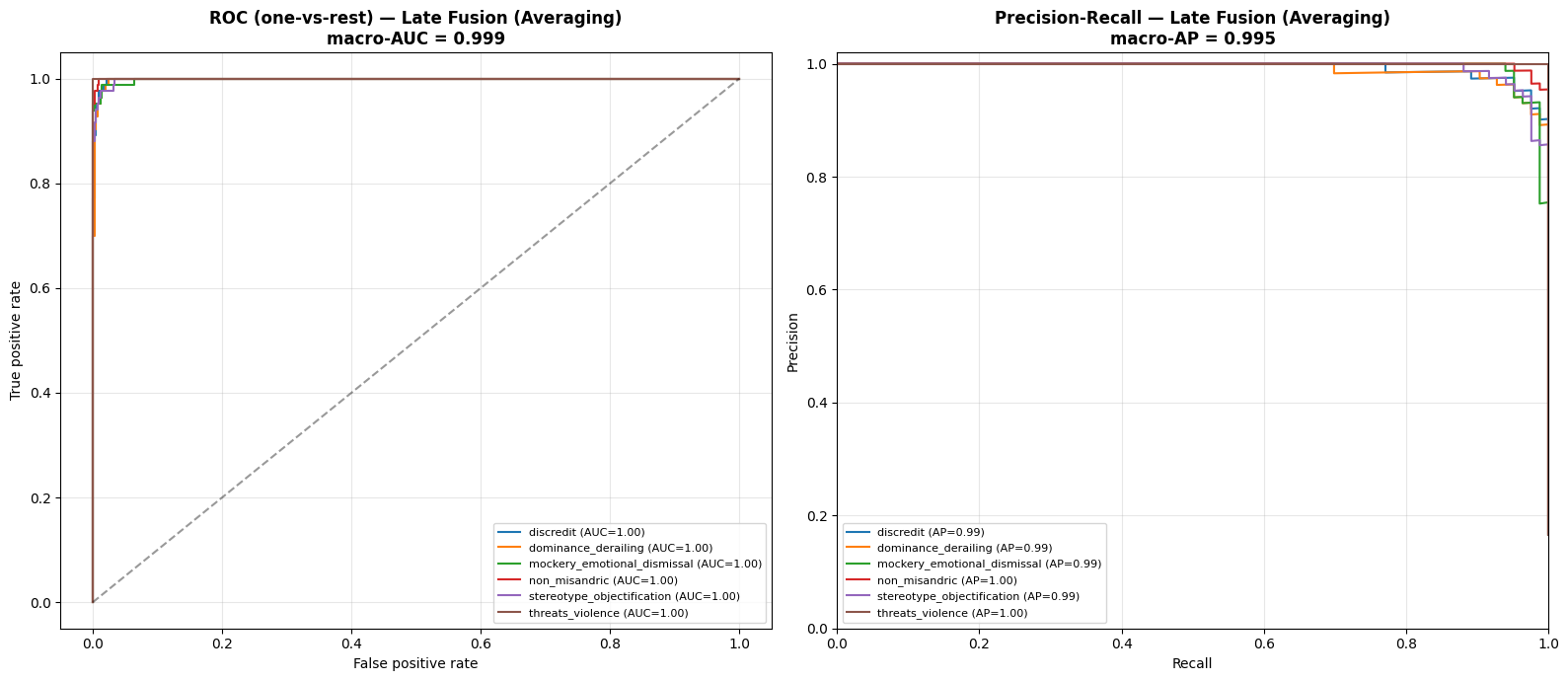}}
\caption{ROC and precision-recall curves, Late Fusion (Averaging). Macro ROC-AUC = 0.999, macro AP = 0.995.}
\label{fig:rocpr}
\end{figure}

Fig.~\ref{fig:rocpr} reports one-versus-rest ROC and precision-recall curves. Late Fusion achieves the highest macro ROC-AUC (0.999) and average precision (0.995) of any model, confirming fusion improves ranking quality, not just the hard decision boundary.

\subsection{Explainability}

To verify that Late Fusion's decision boundary reflects the taxonomy rather than a spurious correlation, we compute Shapley additive explanations \cite{b10} over individual test predictions:
\begin{equation}
\phi_i = \sum_{S \subseteq F \setminus \{i\}} \frac{|S|!\,(|F|-|S|-1)!}{|F|!}\Big[v(S\cup\{i\}) - v(S)\Big]
\label{eq:shap}
\end{equation}
where $F$ is the token set, $S$ ranges over subsets excluding token $i$, and $v(S)$ is the model output restricted to $S$; this is the unique attribution satisfying local accuracy, missingness, and consistency \cite{b10}. A LIME cross-check \cite{b11} surfaced the same top tokens, so we report only SHAP here.

\begin{figure}[htbp]
\centerline{\includegraphics[width=0.85\linewidth]{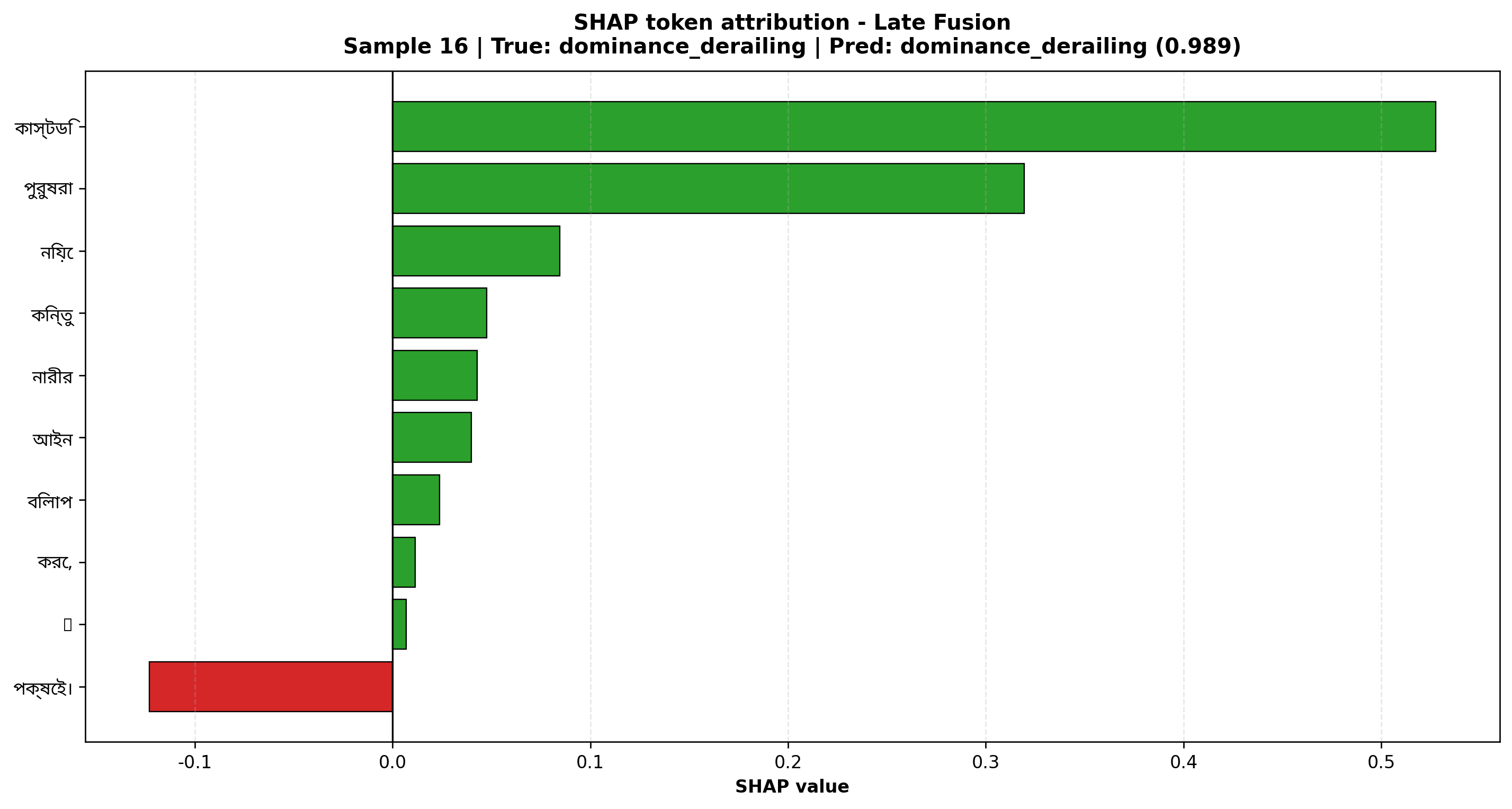}}
\caption{SHAP token attribution for Late Fusion on a held-out dominance\_derailing comment. Green pushes toward the predicted class, red away from it.}
\label{fig:shap}
\end{figure}

Fig.~\ref{fig:shap} shows the attribution for a held-out \texttt{dominance\_derailing} comment predicted with 0.989 confidence. \textit{Kastodi} (custody) carries the largest positive contribution, followed by \textit{purushra} (men); \textit{kastodi} is also the single highest log-odds word for this class ($+10.63$), converging evidence that the classifier's decision boundary tracks the taxonomy's defining vocabulary.

We are careful not to overstate this convergence. SHAP, LIME, and the log-odds statistics are all computed on the same synthetic corpus, so their agreement shows internal consistency with the data as generated, not that the model has acquired genuine semantic understanding of misandry or would generalize to real-world text. Words such as \textit{kastodi} or \textit{jele} may be predictive partly because GPT-4.1 leans on a narrower set of framing devices per category than authentic discourse would, so the classifier may be partly exploiting generation-specific lexical shortcuts rather than, or alongside, genuine semantic indicators; distinguishing the two requires testing on independently sourced real-world text, left to future work.

\section{Discussion}

The dominant confusion is between the non-misandric control class and the two most conversational misandric classes, \texttt{mockery\_emotional\_dismissal} and \texttt{dominance\_derailing}. We read this as a taxonomy property, not a modeling failure: these classes layer dismissive intent onto ordinary discussion of men's issues, so their vocabulary overlaps with the pro-equality register defining \texttt{non\_misandric} (\textit{manoshik sasthya}, mental health, or \textit{kastodi}, custody, appear in either frame).

The 4.8 point classical-transformer gap is smaller than typically reported for bag-of-words versus fine-tuned transformers on English hate speech benchmarks \cite{b22}, plausibly because character n-grams approximate sub-word morphology without a learned embedding space; a linear SVM over character TF-IDF is a credible fallback when GPU inference is unavailable, at a cost of roughly five accuracy points.

Because Ben-Misandry v1 is entirely synthetic, its word distributions should not be read as ground truth: content-filter-constrained generation favors rhetorical or institutional threat phrasing over the extreme unmoderated register, likely making \texttt{threats\_violence} somewhat conservative relative to the real-world tail. More fundamentally, GPT-4.1 is trained toward alignment objectives that discourage genuinely hateful output and is more extensively optimized for English than Bengali, which can bias generated misandry toward whatever register a safety-aligned, English-centric model treats as an acceptable stand-in for the real thing. The 96.6\% accuracy in Section V should therefore be read as evidence the pipelines separate GPT-4.1's six taxonomy-conditioned writing styles reliably, and only tentatively as evidence they detect real-world Bengali misandry; the 500-row human check (Section III-B) supports label-taxonomy correspondence within the synthetic distribution but is not a substitute for real-world evaluation.

\section{Conclusion and Future Work}

We introduced Ben-Misandry, the first Bengali dataset targeting misandry detection, benchmarked with classical and transformer methodology. A linear SVM over character-level TF-IDF reaches 91.8\% test accuracy at negligible cost, while a late fusion ensemble of BanglaBERT and XLM-RoBERTa reaches 96.6\% accuracy and macro ROC-AUC 0.999, nominally ahead of both single encoders and early fusion, though we have not established statistical significance nor directly measured error correlation between encoders. Shapley attributions track the same vocabulary that independently characterizes each class in the raw corpus, which we read as internal consistency rather than confirmed evidence against lexical-shortcut learning. Given the dataset's synthetic, English-centric-model origin and our single-rater 500-row human check, we present these results as a feasibility demonstration rather than a validated real-world benchmark.

Future work priorities: a fully overlapping multi-annotator pass reporting Cohen's $\kappa$ and Krippendorff's $\alpha$; a real, ethically sourced seed corpus to augment the taxonomy and evaluate on authentic text; direct error-correlation measurement and paired significance testing; and multi-label annotation, since a comment can carry more than one misandric category. We hope Ben-Misandry can enable cross-corpus experiments comparing misogyny and misandry classifiers, toward moderation that treats gender based hate as genuinely bidirectional.

\section*{Acknowledgment}
The authors thank the open-source maintainers of the BanglaBERT, XLM-RoBERTa, and BUET Bengali normalizer projects, whose public releases made this work possible.

\end{document}